\documentclass[10pt, a4paper, numbers=noenddot]{article}
\usepackage{lrec}
\usepackage{multibib}
\newcites{languageresource}{Language Resources}
\usepackage{tabularx}
\usepackage{soul}
\usepackage{times}
\usepackage{latexsym}
\usepackage{url}

\usepackage{algorithm,algpseudocode}
\usepackage{graphicx}
\usepackage{multirow}

\usepackage{amsthm}
\theoremstyle{definition}
\newtheorem{exmp}{Example}[section]
\usepackage{xcolor}

\usepackage{epstopdf}
\usepackage[utf8]{inputenc}

\usepackage[bookmarks=true,
            bookmarksopen=true,
            bookmarksnumbered=true,
            colorlinks=true,
            citecolor= magenta,
            urlcolor  = blue,
            linkcolor=blue                 
            ]
            {hyperref}

\usepackage{xstring}

\def\thesection{\arabic{section}}
\def\thesubsection{\arabic{section}.\arabic{subsection}}

\title{Automatic Compilation of  Resources for Academic Writing \\ and Evaluating with 
 Informal Word Identification and Paraphrasing System}

\name{Seid Muhie Yimam$^1$, Gopalakrishnan Venkatesh$^2$, John Sie Yuen Lee$^3$ and Chris Biemann$^1$  }

\address{Universität Hamburg, Germany$^1$, International Institute of Information Technology,  Bangalore, India$^2$, \\ City University of Hong Kong, Hong Kong SAR$^3$ \\
\tt{yimam@informatik.uni-hamburg.de }}

\abstract{
 We present the first approach to automatically building resources for academic writing. The aim is to build a writing aid system that automatically edits a text so that it better adheres to the academic style of writing. On top of existing academic resources, such as the Corpus of Contemporary American English (COCA) academic Word List, the New Academic Word List, and the Academic Collocation List, we also explore how to dynamically build such resources that would be used to automatically identify informal or non-academic words or phrases. The resources are compiled using different generic approaches that can be extended for different domains and languages. We describe the evaluation of resources with a system implementation. The system consists of an informal word identification (IWI), academic candidate paraphrase generation, and paraphrase ranking components. To generate candidates and rank them in context, we have used the PPDB and WordNet paraphrase resources. We use the \emph{Concepts in Context} (CoInCO) "All-Words" lexical substitution dataset both for the informal word identification and paraphrase generation experiments. Our informal word identification component achieves an F-1 score of 82\%, significantly outperforming a stratified classifier baseline. The main contribution of this work is a \emph{domain-independent} methodology to build targeted resources for writing aids.
 \\ \newline  }

\begin{document}

\maketitleabstract

\section{Introduction}

We present the first approach to building resources for an academic writing aid system automatically. Academic writing aid systems help in automatically editing a text so that it better adheres to the academic style of writing, particularly by choosing a better academic word in a given domain. In the context of academic paraphrasing tasks, the resources are mainly words or phrases, that are more appropriate to use in an academic writing style. Moreover, the academic resources might vary from domain to domain as some words or phrases are extensively used in one domain over the other. 

%We construct the resources that aid in academic writing and use them in an academic writing aid.
The first step in building an academic writing aid tool is to collect resources that determines whether a given phrase follows the style of writing in academia. This involves analyzing a given sentence and determining if the lexemes of the sentences are well-selected academic words and phrases or not. 

To evaluate the resources compiled, we have to build a system,
 analogous to the lexical substitution and text simplification tasks, for example, \cite{szarvas-etal-2013-supervised,stajner-saggion-2018-data}, that consists of informal word identification, academic candidate generation, and candidate paraphrase ranking components (see Figure \ref{fig:net_trigrams}).
While it is possible to follow the same approaches as the lexical substitution and text simplification approaches for academic text rewriting tasks, the main challenge for the academic paraphrasing task is the collection of resources for academic texts. 

The following are the main objectives of building academic resources:
\begin{enumerate}
% \item Investigate how existing academic resources (word and phrase lists) are built. 

 \item Identify suitable academic and non-academic datasets that are to be used to build academic resources.

\item Design a generic, \emph{domain-independent}, approach to extract academic resources.

\item Evaluate the quality of the collected resources and use these resources for informal word identification (IWI) and academic paraphrasing systems.
 
\end{enumerate}

The informal word identification (IWI) component automatically identifies informal words (see Section \ref{informalded}) that are going to be replaced with academic paraphrases. The candidate generation and ranking component determine the best academic candidate paraphrase to replace the informal words. 

The ultimate goal of this research work is to integrate the informal word identification, candidate generation, and paraphrase ranking components into writing aid tools, for example to word processors or text composing software like latex packages, to automatically assist users in academic text composing.

In this work, we have targeted the following research questions 1)\textbf{ How to build academic resources (words or phrases), which are used to replace informal or less academic expressions in academic texts?} 2) \textbf{How to build a system that can be used to evaluate the collected resources?}

%\noindent \textbf{RQ2}: How to build a system that will automatically detect less academic words and phrases (informal word identification - IWI) and propose more formal or academic replacements (academic text paraphrasing)?

In Section \ref{relatedworks}, a brief review of related works is presented. In Section \ref{buildresource}, we discuss how to build academic resources using reference corpora and evaluate the quality of the resource. In Section \ref{academicsystem}, we present the approaches that are used to build an informal word identification and paraphrasing system for academic rewriting. Setups of the academic paraphrasing systems and the experimental results are discussed in Section \ref{experiments}. Analysis of system results and conclusion of the research are presented in Section \ref{analysis} and Section \ref{conclusion} respectively.

\section{Previous Work}
\label{relatedworks}
In this section, we review previous work in lexical substitution, a closely related task, and discuss how the academic text rewriting system potentially differs.

In essence, our system is similar to lexical substitution (LS) and text simplification tasks, in such a way that both focus on the rewriting of an original text towards a given goal. Lexical substitution system mainly focuses on rewriting texts by replacing some of the words or phrases without altering the original meaning \cite{szarvas-etal-2013-supervised,stajner-saggion-2018-data}. The work by \newcite{guo-etal-2018-dynamic} targeted text simplification based on the sequence-to-sequence deep neural network model, where its entailment and paraphrasing capabilities are improved via multi-task learning.

While the complex word identification (CWI) task focuses on identifying lexical units that pose difficulties to understand the sentence \cite{R17-1104,yimam:2017:ijcnlp,yimam-etal-2018-report,paetzold-specia-2016-semeval}, our informal word identification (IWI) component focuses on identifying words that are not fitting or adhering to the academic style of writing.

The work by \newcite{riedl-etal-2014-lexical} focuses on the lexical substitution task, particularly for medical documents. They have relied on Distributional Thesaurus (DT), computed on medical texts to generate synonyms for target words.

Existing resources for academic writing are limited to a precompiled list of words such as the Corpus Of Contemporary American English (COCA) \cite{10.1093/applin/amt015} and the New Academic Word List 1.0 (NAWL) \cite{Charles2019} vocabulary lists. Regarding phrases (multi-word expressions) for academic writing, the only available resources are the academic bi-grams compiled by Pearson\footnote{Academic collocation list: \url{https://pearsonpte.com/organizations/resea}}.

However, these resources are 1) limited to a certain domain and target writers (mostly L2 learners and students), 2) their vocabulary is fixed, thus requiring manual work for an extension, and 3) the resources are limited to uni-gram and bi-gram lists. In this work, we build academic resources that are more generic, which can be built from existing reference corpora. In addition to uni-gram and bi-gram resources, we also design a system that can produce resources up to a length of four words (quad-grams).

As far as we know, the only system available to academic writing is the work of \newcite{lee-etal-2018-assisted}, which addresses a different aspect, which is a sentence restructuring based on nominalizing verbal expressions.

%\section{Definitions}
%In the following subsections, we provide a formal definition of academic/informal words and phrases as well as how we build the dataset for each component in the pipeline for academic text rewriting.
\section{Building Academic Resources}
In this section, we will first discuss the existing academic resources, how they are built and their limitations. Then, we will present our approach that describes the process of building academic resources from different reference corpora. Finally, we will discuss the quality of the collected resources against two evaluation measures, namely comparing with the existing resources and manually evaluating the academic fitness of resources.   
\label{buildresource}
\subsection{Existing Resources for Academic Writing} 
In this subsection, we will present the existing academic word lists and phrases, which will be used to evaluate the quality of the dataset we build from reference corpora.
\label{existinresources}
\subsubsection{Academic Vocabulary}

There are some efforts in building a list of vocabularies or words for academic writing. Some of them are created by analyzing text from academic writing corpora such as journals, theses works, and essays. One such resource is the Corpus Of Contemporary American English (COCA) \cite{10.1093/applin/amt015} vocabulary list, which contains about 3,000 words (in lemmas) that are derived from a 120 million word sub-corpus of the 560 million words. Similarly, the New Academic Word List 1.0 (NAWL) \cite{Charles2019} was also built in the same way as the COCA list as a reference resource for second language learners of English, which is selected from an academic corpus of 288 million words.

\subsubsection{Academic Phrases}

Academic phrases are a list of collocated words (multi-word expressions), which are mostly used for academic writing. The list from \newcite{article} comprises of 2,468 bi-gram collocations. The list is compiled from the written curricular component of the Pearson International Corpus of Academic English (PICAE) comprising of over 25 million words. However, the academic phrases, like the academic word lists, are mostly used as a guideline (study material) to practice academic writing.

\subsection{Academic and Non-Academic Reference Corpora}

The existing resources that are presented in Section \ref{existinresources} are prepared mostly as references or study guidelines for academic writers. However, to build automatic writing support, it is required to have more comprehensive and larger resources that can also be updated dynamically. In addition to single word and bi-gram lists, it would be also beneficial if the resource includes longer sequences of words. Hence, we have further extended the academic phrase list that includes up to four-gram phrases. The resource helps the academic paraphrasing or rewriting system in 1) identifying words or phrases in a text that are less academic and 2) providing alternative academic words or phrases that are more relevant to the contexts presented.

\begin{figure*}[]
\centering
\includegraphics[width=\textwidth]{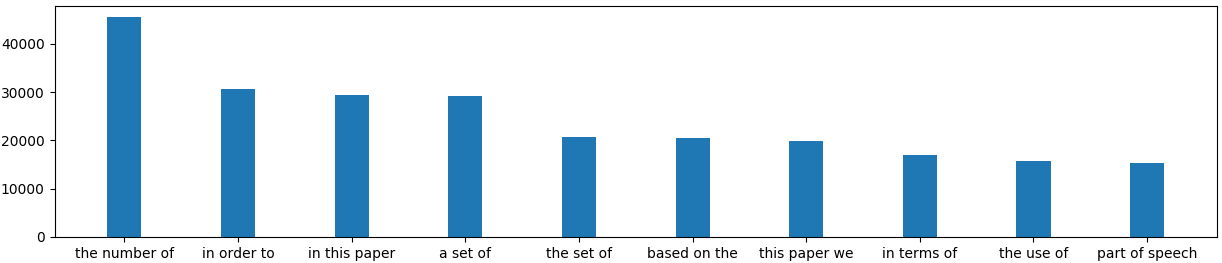}
\caption{Frequencies of the highest occurring tri-grams collected from the reference corpora based on our approach.}
\label{fig:net_trigrams}
\end{figure*}

To this end, we have compiled a list of academic phrases that are extracted from the ACL Anthology Reference Corpus (ACLAC) \cite{L08-1005}. This corpus contains 22,878 scholarly publications (articles) about Computational Linguistics. To understand the syntactic difference of an academic corpus from a non-academic corpus, we have used the Amazon Review Full Score Dataset \cite{Zhang:2015:CCN:2969239.2969312} as our non-academic reference. The non-academic dataset is constructed by randomly taking 600,000 training samples and 130,000 testing samples for each review score from 1 to 5 \cite{Zhang:2015:CCN:2969239.2969312}. In this paper, a review refers to the review text from the training sample.

\begin{table}[]
\centering
\begin{tabular}{|l|r|r|}
\hline
Resource & Size & Coverage (\%)\\ 
 \hline
COCA & 3,015 & 95.39 \\
NAWL & 963 & \textbf{99.90} \\
Academic phrases & 2,468 & 79.34\\
\hline
\end{tabular}
\caption{Coverage of the existing resources for academic writing in our reference ACLAC corpus.}
\label{table:corpus_unigrams}
\end{table}

The above two corpora can be considered to be a good fit as it shows a high match with the existing academic vocabulary or phrase list, as shown in Table \ref{table:corpus_unigrams}. From Table \ref{table:corpus_unigrams}, we can see that 95\% of the academic words from COCA and 99.90\% of the academic words from NAWL are represented in the ACLAC corpus. Similarly, around 80\% of the bi-grams from the academic phrases (PICAE) are contained in the ACLAC corpus.

\subsection{Approach to Build the New Academic Resource}
\label{resourceapparoch}
On analyzing the corpora, we noticed that the non-academic corpus is much larger (in terms of the number of words) than the academic corpus. Therefore, we downsampled the non-academic text (to have comparable resources in terms of size) and ensured that the total number of words in both of the corpora are comparable. As a part of the pre-processing step, we clean the corpus (removing special characters) and lower case each word. We have considered a total of 991,798 reviews, which results in 75,184,498 tokens.

Using the NLTK's\footnote{\url{https://www.nltk.org/}} Bi-, Tri- and Quad-Gram multi-word expression finder, we have extracted phrases from the two corpora (ACLAC and Amazon Review Full Score Dataset) and also compute the frequency distributions of these phrases across both the corpora as it can bee seen in Figure~\ref{fig:net_trigrams}. The phrases extracted from both corpora can be used to assess naively the distribution across the two domains. 

However, we have followed two different widely adopted approaches to extract representative phrases in a corpus, which is specifically known as keyphrases.
The first approach is called Term Frequency-Inverse Document Frequency (TF-IDF), which is one of the most important statistics that show the relative importance of a term in a document in comparison to the corpus. The importance increases proportionally to the number of times a word appears in the document while its weight is lowered when the term occurs in many documents. We used the scikit-learn\footnote{\url{https://scikit-learn.org/}} implementation of TF-IDF to compute the scores of the different n-grams and thereby select the phrases that have maximum TF-IDF scores as keyphrases. In the ACLAC corpus, we have considered an article as one document while for the Amazon Review dataset, a review is considered as a single document.

In the second approach, we explore keyphrase extraction techniques based on part-of-speech sequences. We have employed \emph{EmbedRank}, an unsupervised keyphrase extraction tool trained with sentence embeddings \cite{bennani-smires-etal-2018-simple}. We consider only those phrases that consist of zero or more adjectives followed by one or multiple nouns \cite{Wan:2008:SDK:1620163.1620205}. While using the official implementation\footnote{\url{https://github.com/swisscom/ai-research-keyphrase-extraction}}, we also explored the possibility of using the \emph{Spacy}\footnote{\url{https://spacy.io/}} POS tagger for keyphrase extraction in our corpora, which has a permissive license to redistribute our resource generation system as an open-source project.

As per the heuristic approach followed in the COCA word list compilation, we only retain those phrases that occur at least 50\% more frequently in the academic portion of the corpora than would otherwise be expected. In other words, the ratio of the academic frequency of a term (in the ACLAC dataset) to the non-academic frequency (in the Amazon Review Full Score Dataset) should be 1.50 or higher \cite{10.1093/applin/amt015}. Using a similar approach, we have also created the non-academic resources, which are also used to evaluate the quality of the academic resources in the human evaluation experiment (cf. Section \ref{humaneval})

\subsection{Newly Collected Academic Resources}
Based on the two keyphrase extraction approaches discussed in Section \ref{resourceapparoch} (TF-IDF and EmbedRank based keyphrase extractions), we have compiled a total of 6,836 academic phrases (5,275 from EmbedRank and 1,900 from the TF-IDF approach). From Table \ref{table:collected_resource}, we can see that most of the academic keyphrases are extracted using the EmbedRank approach. %In the future, it would be nice to extend the EmbedRank approach to extract keyphrases beyond the adjective and noun POS tag patterns, especially to cover verbs used in academic contexts.
%As we can see from Table \ref{table:collected_resource}, more bigrams are extracted from EmbedRank than unigrams. %This due to the pattern

\begin{table}[]
\centering
\resizebox{1.0\linewidth}{!}{
\begin{tabular}{|l|r|r|r|r|}
\hline
\multicolumn{5}{|c|}{Newly Collected resources}\\
\hline
Approach & Uni-gram & Bi-gram & Tri-gram & Quad-gram\\
\hline
EmbedRank & 1,267 & 3,848 & 156 & 4 \\
TF-IDF & 1,090 & 690 & 109 & 11 \\
\hline
\multicolumn{5}{|c|}{From Existing Resources}\\
\hline
COCA & 3,016 & 0 & 0 & 0 \\
NAWAL & 960 & 0 & 0 & 0 \\
PICAE & 0 & 2,468 & 0 & 0\\
\hline
\end{tabular}
}
\caption{Academic word and phrases lists from the existing as well as from newly collected resources. %The row \emph{AC} stands for Academic collocation. 
}
\label{table:collected_resource}
\end{table}
\subsection{Manual Evaluation of Resources}
\label{humaneval}
From the automatically compiled list of resources (words and phrases), we have randomly sampled 520 words and phrases comprising of \textbf{155 uni-grams}, \textbf{100 bi-grams} and \textbf{5 tri-grams} from each of the compiled academic and non-academic phrase list. We then distributed the word and phrase lists to a total of 9 annotators (Ph.D. and postdoctoral researchers)  and requested the participants to label each entry as academic or non-academic. The sampled words and phrases are evaluated by two sets of annotators and the annotators were able to label the entries with an inter-annotator agreement of \textbf{68.22\%}. %While we can consider that the two generic and automatic approaches to extract academic phrases works reasonably well, a more rigorous approach should be adopted to improve the quality of the resources. %Besides the accuracy, we have also evaluated the agreements of two annotators on the same entry using Spearman correlation.

\subsection{Results and Discussions on the Collected Resources}

While analyzing the COCA list, we noticed that it contains a few stop words such as {\textbf{both}} and {\textbf{above}}. Hence, while relying on TF-IDF, we have considered extracting academic resources in different scenarios. First, we remove stop words as a part of the preprocessing step and in the second approach we have used the whole corpus as it is.% As it can be seen from Table \ref{table:human_evaluation_unigrams}, removing the stop words results in a lower coverage of academic resources in the COCA and NAWL lists.

The system proposed by us relies on the relative frequencies in the reference corpora which can be computed independently of the language used. Thus the compilation of such an academic resource (through keyphrase extraction) can be considered language agnostic.

While performing the human evaluation, the annotators were asked to classify whether the given phrase is academic or not. The evaluation would have been more rigorous if they had to classify the phrases given the context in which the term had occurred. The annotators have at times labeled an entry as both academic and non-academic. Consider the word \textbf{attention}, it was used both in an academic (ACLAC) and non-academic (Amazon Review Full Score Dataset) context, for example as "LSTM with \textit{\color{blue}attention}" and "the kid's \textit{\color{blue}attention} to the game" respectively.

%\vspace{0.5em}
%\fbox{\begin{minipage}{18em}
%        \begin{exmp}
%            \small
%            \label{ex:exampleatten}
%            \hfill \break
%            \hfill \break
%            \textbf{Academic}: The {\color{blue} \textbf{attention}} vector is then created by averaging weights over all input sentences.
%            \newline \hfill \break
%            \textbf{Non-academic}: It doesn't keep the kids {\color{blue}\textbf{attention}} because the kid wants to sit and listen to a bunch talking.
%        \end{exmp}
%    \end{minipage}
%}
%\vspace{0.5em}

\section{Evaluating the Resources for Academic Rewriting System}
\label{academicsystem}
\subsection{Academic Words}
We define a word as \textbf{academic} or \textbf{formal} if it is in one of the following lists of academic phrases 1) keyphrases (up to four-grams) compiled by our system (cf. Section \ref{resourceapparoch} -- comprises of 6,836 phrases) 2) the COCA list \cite{Davies:2012:Online} 3) the New Academic Word List \cite{Charles2019}\footnote{ \url{http://www.newgeneralservicelist.org/nawl-new-academic-word-list}}. %Otherwise, it is termed as \textbf{non-academic} or \textbf{informal} if it is not in one of the above lists.

Some example academic words are shown in Table~\ref{tab:academic}.
The academic word lists are also extended to phrases or multi-word expressions. Pearson has published a set of academic bi-grams\footnote{Academic collocation list: \url{https://pearsonpte.com/organizations/resea}}. Words like \emph{best}, \emph{almost}, and \emph{way} are not by themselves \emph{academic}, but they can be combined with other words to form academic expressions such as \emph{best described}, \emph{almost identical}, and \emph{appropriate way}.

\subsection{Informal Words}

\label{informalded}
The naive approach is to attempt to rewrite every non-academic word, using our definition above. That is a misplaced goal, however, since even the average document in the BAWE corpus \cite{alsop2009issues} contains a considerable number of words outside the list, including function words and other words commonly used in all English documents.

We define a word as \textbf{informal} if it is a non-academic term that can be paraphrased by an academic term. If the term is academic, or it is non-academic but does not have an academic paraphrase, it is termed as \textbf{formal}. %In other words, no academic word is informal; some of the non-academic words are informal, while others are not because they cannot be paraphrased. 

\subsection{Architecture}

\begin{figure*}
 \centering
 \includegraphics[width=0.88\textwidth]{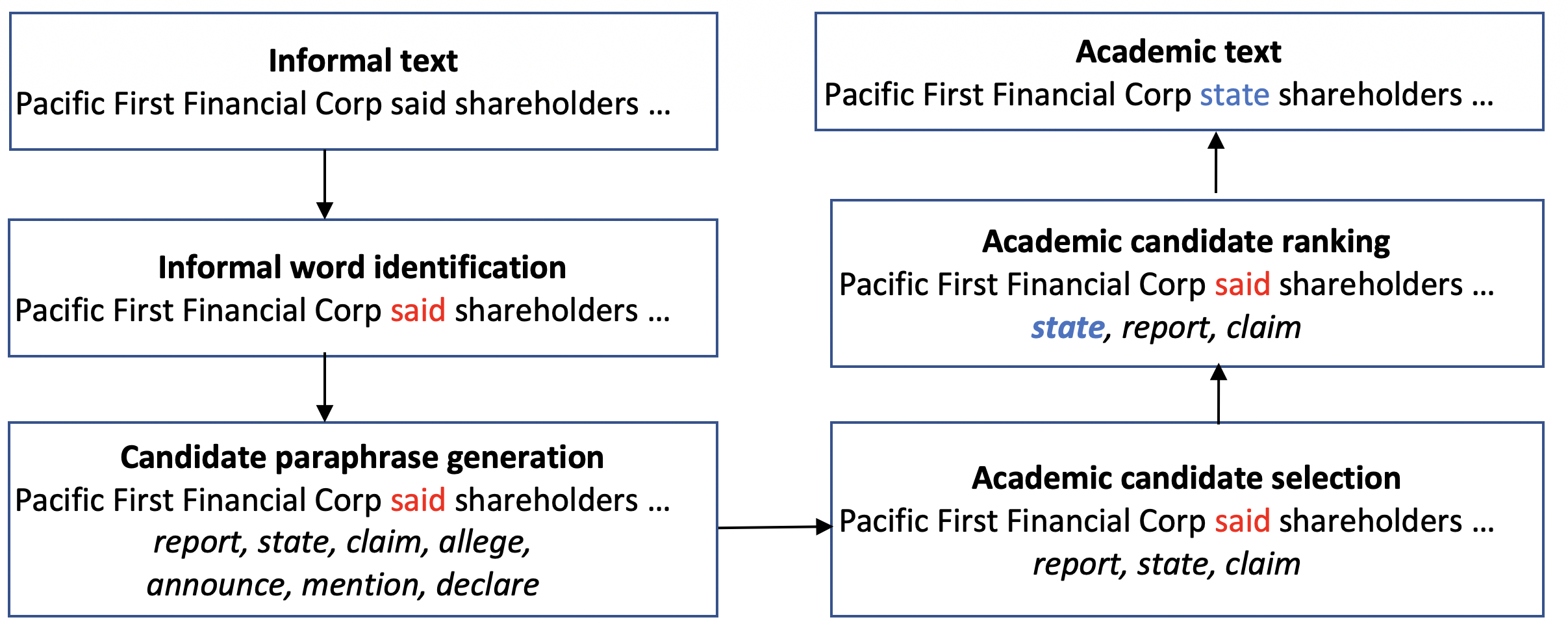}
 \caption{Architecture of the system.}
 \label{fig:architecture}
\end{figure*}
As shown in Figure \ref{fig:architecture}, our proposed system consists of four components, which is analogous to the lexical simplification systems \cite{Paetzold:2017:SLS:3207692.3207704}. The components of our system constituted informal word identification (IWI), paraphrase generation, candidate selection, and paraphrase ranking.

\subsubsection{Informal Word Identification}

The informal word identification (IWI) component identifies each word as \emph{informal}, or not. The system attempts to paraphrase only the informal words in the rest of the pipeline.

Similar to CWI \cite{R17-1104,yimam:2017:ijcnlp,yimam-etal-2018-report,paetzold-specia-2016-semeval}, IWI is more accurate when placed in context. The word \emph{big}, for example, may need to be paraphrased to \emph{major} in the context of "\emph{This article makes two big contributions.}" It should not be paraphrased, however, when it is part of the expression \emph{big data}.

\subsubsection{Paraphrase Generation, Selection, and Ranking}

Given an informal word, this step generates a list of substitution candidates. While there are different approaches to generate candidates for target words, such as using existing paraphrase resources like WordNet and Distributional thesaurus (see \newcite{yimam-EtAl:2016:MWE}), we depend solely on the CoInCo \cite{kremer-etal-2014-substitutes}, WordNet \cite{Miller:1995:WLD:219717.219748}, and the paraphrase database (PPDB) \cite{P15-2070} resources to generate candidates. 

Once the candidates are generated, all of the candidates, which must be academic words are retained for the paraphrase ranking component. Given a list of academic substitution candidates, the paraphrase ranking component finds the one that fits best in the context. The detailed approach is presented in Section \ref{data}.

\begin{table}[!t]
    \centering
%    \resizebox{1.0\linewidth}{!}{
        \begin{tabular}{|l|l|}
            \hline
            Academic words & report, state, claim... \\
            \hline
            Non-academic words & say, declare, mention, allege... \\
            \hline
        \end{tabular}
%    }
    \caption{\label{tab:academic}Example of academic and non-academic words based on our academic resources.}
\end{table}

\subsection{Datasets for IWI and the Paraphrasing Components}

\label{data}
For this evaluation, we derive our dataset from a lexical substitution dataset called the Concepts in Context (CoInCo) \cite{kremer-etal-2014-substitutes}. The CoInCo dataset is an \emph{All-Words} lexical substitution dataset, where all words that could be substituted are manually annotated. The corpus is sampled from newswire and fiction genres of the Manually Annotated Sub-Corpus (MASC) corpus\footnote{\url{http://www.anc.org/data/masc/}}. While the targets (words that are going to be substituted) are used to build the informal word identification dataset, the candidates are further processed to perform the academic paraphrase ranking task.% On the other hand, in the \newcite{yimam-EtAl:2016:MWE} dataset, 5 target words are randomly selected (verb, noun, adjective, and adverbs) where candidates are automatically generated from paraphrase resources but they are manually ranked for their fitness in context. We split both datasets into a training set and a test set.

A total of 1,608 training and 866 test sentences are compiled out of 2474 sentences from the CoInCo dataset. %Only 20 sentences do not provide at least one informal and one formal target/candidate in their set.
Statistics on the IWI dataset are shown in Table~\ref{tab:iwi}.

\subsubsection{Building the IWI Dataset}

We automatically generated an IWI dataset from CoInCo as follows. For each non-academic target word, we determine if its substitution candidates include at least one academic word. If so, it is labeled as \textbf{informal}; otherwise, it is labeled as \textbf{formal}. All academic target words and all words without substitution candidates are labeled as \textbf{formal}. An example is given in Example~\ref{ex:example} and Table~\ref{tab:academic}.

%In this example, let's show one target word whose substitution candidates are all non-academic, to make our definition clear 

\begin{table}[!t]
\centering
\resizebox{1.0\linewidth}{!}{
\begin{tabular}{|l|l|}
\hline
CoInCo annotation &    Pacific First Financial Corp said[paraphrases: report, state, detect] shareholders \\    
IWI dataset & Pacific[N] First[N] Financial[N] Corp[N] said[Y] shareholders \\
\hline
\end{tabular}
}
\caption{\label{tab:iwi_example} Transformation of the CoInCo dataset into IWI dataset, with respect to the academic word list in Table~\ref{tab:academic}}
\end{table}
\vspace{0.5em}
\fbox{\begin{minipage}{18em}
        \begin{exmp}
            \small
            \label{ex:example}
            \hfill \break
            \textbf{Sentence}: Pacific First Financial Corp said shareholders ...
            \hfill \break
            \textbf{CoInCo annotation}: \underline{Target word}: {\color{red}said}. \underline{Paraphrases}: {\color{blue}report, state, claim}, {\bf{allege, announce, mention, declare}}
            \hfill \break
            \textbf{IWI dataset} ({\color{red}[I]}--informal, [F]--formal): Pacific[F] First[F] Financial[F] Corp[F] said{\color{red}[I]} shareholders[N]
        \end{exmp}
    \end{minipage}
}
\vspace{0.5em}

%Note: it's possible to expand the IWI this way: if the target word is academic and at least one of its substitution candidates is non-academic, we can pretend the non-academic candidate to be the target word (therefore an informal word)

\begin{table}[!t]
    \centering
%    \resizebox{1.0\linewidth}{!}{
        \begin{tabular}{|l|r|r|r|r|}
            \hline
            \hline
            Dataset & \multicolumn{2}{|c|}{\# Tokens} & \multicolumn{2}{|c|}{\#Types} \\
            \cline{2-5}
            & I & F & I & F \\
            \hline
            IWI training & 6,783 & 3,358 & 2,266 & 1,509 \\
            IWI test & 3,666 & 1,822 & 1,577 & 994 \\
            \hline
        \end{tabular}
%    }
    \caption{\label{tab:iwi} Statistics on the IWI dataset. \emph{\#Tokens} shows the total number of tokens (formal (\emph{F}) and informal (\emph{I})) while \emph{\#Types} shows the unique occurrences of tokens in the IWI training and test sets. \emph{I} stands for informal and \emph{F} for formal tokens and types resp.}
\end{table}

%Nathans_Bylichka.txt. s-r886 Now leave , before I call the rats on you . ” call    v    1
%Nathans_Bylichka.txt    s-r886    Now leave , before I call the rats on you . ”     the    DT    0 )

\subsubsection{Paraphrase Candidates}
\label{candidates}
To generate \textbf{non-academic} to \textbf{academic} word pairs for paraphrasing, we used the paraphrases (word pairs) in CoInCo, WordNet, and PDPB as the starting point. 

For the CoInCo dataset, we have only included those word pairs where: 1) the target word is non-academic, 2) the substitution candidate is academic, 3) the target word has a higher word frequency than the substitute candidate in our academic resources. %(around 21K unique entries).
Since the academic resource is not exhaustive, some proper academic terms may be mistakenly considered as \textbf{non-academic}. This requirement aims to prevent these words from being substituted.

For example, from the sentence in Example~\ref{ex:example}, we obtained the word pairs {\color{red}say}:{\color{blue}report}, {\color{red}say}:{\color{blue}state}, and {\color{red}say}:{\color{blue}claim}. We have collected a total of 23,476 word pairs from the CoInCo Training Set.

The dataset is prepared with 4 candidates for each informal target, where 2 candidates are academic and 2 candidates are non-academic. When we do not have appropriate candidates we extract further candidates from WordNet \cite{Miller:1995:WLD:219717.219748} and PPDB \cite{P15-2070}. Table \ref{tab:corpus} shows the statistics of target words extracted from the CoInCo dataset, where 59\% of the informal words have possible candidate paraphrases.

% For the PPDB \cite{P15-2070} paraphrase candidate generation, we follow a similar approach as the CoInCo dataset. For each paraphrase pairs in the PPDB ( targets and candidates), we retain all targets that are non-academic and have at least one academic candidate. We also make sure that the target entry has a higher frequency than the candidate in the general COCA word lists. A total of 525,775 word pairs are collected from PPDB that are used to generate candidate paraphrases.

\subsection{Academic Paraphrase Corpus}

In general, any existing paraphrase or lexical substitution corpus can be converted into an academic paraphrase corpus with the following steps:

    \noindent 1) Discard all academic target words since they do not need to be paraphrased.\\
    \noindent 2) Remove all non-academic substitution candidates for the remaining (non-academic) target words.\\
    \noindent If no candidate is left after step (2), also remove that target word.

%We used the above procedure to create the \emph{Yimam Ranking Test Set} from the test set in \newcite{yimam-EtAl:2016:MWE}, and the \emph{CoInCo Ranking Test Set} from the CoInCo test set (see above). Statistics are reported in Table~\ref{tab:corpus}. 

%Overall, 73.15\% of the gold substitution candidates are included among the candidate's target words.%, The gold substitution XXX% 

\begin{table}[!t]
    \centering
%    \resizebox{1.0\linewidth}{!}{
        \begin{tabular}{|r|r|r|}
            \hline
             \multicolumn{2}{|c|}{\# target words} & \multicolumn{1}{|c|}{Paraphrase coverage} \\
            \cline{1-3}
             Original & Our corpus & in (\%) \\
            \hline
             5,480 & 3,250 & 59.30 \\ 
            \hline
        \end{tabular}
%    }
    \caption{\label{tab:corpus} Statistics on our evaluation dataset. The last column shows the percentage of non-academic words in the corpus for which paraphrases can be obtained.}
\end{table}

\subsection{Informal Word Identification Models}

We trained three Support Vector Machine (SVM) classifiers, using Radial Basis Function kernel, from scikit-learn\footnote{\url{https://scikit-learn.org/}} with different feature sets. We use the following features:

    \noindent \noindent \textbf{Word frequency}: We use word frequencies 1) in the Beautiful Data\footnote{\url{https://norvig.com/ngrams/}} which are derived from the Google Web Trillion Word Corpus, 2) in the general COCA list, and 3) in the ACL anthology corpus \cite{L08-1005}.
    
    \noindent \textbf{Word Embedding}: We have used GloVe \cite{pennington2014glove} word embedding to compute the cosine similarity between the word and the sentence\footnote{Embedding for the sentence is calculated by averaging the embedding of words in the sentence}. We also explore the option of using Euclidean distance between the word and the sentence as a feature while training the classifier.
    
    \noindent \textbf{Part of Speech Tag (POS)}: The POS tag of the word obtained from the TreeTagger\footnote{\url{https://www.cis.uni-muenchen.de/~schmid/tools/TreeTagger/}}.
    
    %\noindent \textbf{Problematic Annotation}: A binary feature that is set to \textit{True} if less than two annotators provided a substitute.
    
    \noindent \textbf{Word level features}: We use the word length and the number of vowels as features for training the classifier.

%Compare with features in state-of-the-art CWI systems?

\subsection{Paraphrase Ranking Models}

%We use the existing system described in \newcite{yimam-EtAl:2016:MWE} to rank candidate paraphrases in context.
In order to rank the best candidates for academic rewriting, we have followed the learning-to-rank machine learning approach, where candidates are ranked based on their relevance score. The number of annotators selected the given candidate is considered as a relevance score. The TF-Ranking deep learning model provided by \emph{TensorFlow Ranking}\footnote{\url{https://github.com/tensorflow/ranking}} library \cite{TensorflowRankingKDD2019} is used to build the paraphrase ranking model.

\section{Experiments}
\label{experiments}
\subsection{Informal Word Identification}

We trained the IWI classifier on the CoInCo Train Set using SVM. Similar to most of the CWI evaluation metrics, we evaluate the performance of the system on the following evaluation metrics:

\noindent \textbf{Precision}: The number of correct informal targets, out of all targets proposed by the system.\\
\noindent \textbf{Recall}: The number of correct informal targets, out of all informal words that should be paraphrased.\\
\noindent \textbf{F-Measure}: The harmonic average of precision and recall.

 Table~\ref{tab:iwi_result} shows IWI precision and recalls on the CoInCo Test Set. We use a simple stratified randomization algorithm from scikit-learn as a baseline system. The proposed algorithm (SVM classifier) achieves a better performance overall in the F-Score of 0.8204. As it can be seen in Table~\ref{tab:iwi_result}, the following features work better for the IWI task: frequencies, cosine similarity, and Euclidean distance. 

\begin{table}[!t]
    \centering
    %\resizebox{1.0\linewidth}{!}{
        \begin{tabular}{|l|ccc|}
            \hline
            \hline
            Method & Precision & Recall & F-score \\
            \hline
            Baseline & 0.6679 & 0.6787 & 0.6733 \\
            \hline
            SVM \textit{Fe1} & 0.7584 & \textbf{0.8933} & \textbf{0.8204} \\
         SVM \textit{Fe2} & \textbf{0.7650} & 0.8748 & 0.8162 \\
         SVM \textit{Fe3 } & 0.7552 & 0.8912 & 0.8176 \\
         
            \hline
            \hline
        \end{tabular}
    %}
    \caption{\label{tab:iwi_result} Precision and recall on the informal word identification task. The baseline system has been setup using the Stratified classifier from scikit-learn: The stratified classifier in scikit-learn generates predictions by respecting the training set’s class distribution. Fe1 = (Frequencies, cosine similarity), Fe2 = Fe1 + (Euclidean distance), Fe3 = All features}
\end{table}

\subsection{Academic Paraphrasing}

We evaluate the system performance on automatically generating academic paraphrases and ranking them. Following standard evaluation metrics in lexical simplification, we report on the  \textbf{Mean Reciprocal Rank (MRR)}\footnote{\url{https://en.wikipedia.org/wiki/Mean_reciprocal_rank}} metric.% and \textbf {Normalized Discounted Cumulative Gain (NDCG)} metrics.

The model from TF-Ranking \cite{TensorflowRankingKDD2019} library has been trained to re-rank the candidates on the CoInCo test set. The model was trained using the \emph{Adagrad} optimizer with a learning rate of 0.05. Experiments were performed on various loss functions (\emph{pairwise\_logistic\_loss} and \emph{softmax\_loss}) and different \emph{step}\footnote{Steps are the number of training iterations executed.} (50, 100 and 200) values. Table ~\ref{tab:result_gold} shows the experimental results. %As we have 4 candidates (2 academic and 2 informal), we like to have the two academic candidates to be ranked atop. As it can be seen from the table, the ranking model was able to attain an NDCG@1 score of \textbf{0.7014} and NDCG@2 score of \textbf{0.8814}. 

    %\noindent \textbf{Gold IWI}: the system attempts paraphrasing if and only if the target word is informal (Table~\ref{tab:result_gold}).\\
%    \noindent \textbf{Automatic IWI (Table~\ref{tab:result_gold} lower part)}: The system attempts paraphrasing if the IWI component automatically identifies the word as informal. 

%From Table \ref{tab:result_gold}, we see that using the CoInCo paraphrase generation results in higher performance than using either only PPDB or combination of PPDB and CoInCo. This is because the candidates generated from the PPDB do not fit the context very well. Also, the system ranks well the CoInCo test set better than the Yimam datasets when using only PPDB as a candidate generation. This attributes to the fact that the Yimam dataset contains only a handful of Informal targets per paragraph than the CoInCo (All-words) lexical substitution dataset.
\begin{table}[!t]
    
    \centering
    
    %    \resizebox{1.0\linewidth}{!}{
        \begin{tabular}{|l||c|c|}
            \hline
            \multicolumn{2}{|c|}{Parameters} & Ranking metric\\
            \hline
            Loss & Steps & MRR \\
            \hline
            \multirow{3}{*}{Logistic} & 50 & 0.8861  \\
             & 100 & \textbf{0.8926}      \\
             & 200 & 0.8895     \\
            \hline
            \multirow{3}{*}{Softmax} & 50 & 0.8893  \\
             & 100 & 0.8895     \\
             & 200 & \textbf{0.8914} \\
            
            \hline
        \end{tabular}
    %}
    \caption{\label{tab:result_gold} Academic paraphrasing performance on the CoInCo Test Set using the MRR ranking metric.}
\end{table}

\section{Analysis of Results}
\label{analysis}
For the informal word identification task, our models have a slightly lower precision as our dataset is not balanced (we have more informal words than formal words, as shown in Table \ref{tab:iwi}). %The CoInCo dataset has been compiled from the informal sections of the freely available MASC corpus. is also informal text and we are identifying one target a time in a sentence, the classifier tends to identify the academic targets as informal. 

From an error analysis, we find out that even if the term is academic in general, its usage in the test dataset is inclined to be informal. For example, in the sentence "\textit{It was last February, after the winter {\color{blue} \textit{break}}, that we moved in together.}", \textit{break} is labeled as academic but should be labeled as informal. This issue could be solved by further enhancing the dataset by employing human annotators during the resource compilation process. 

Similarly, some of the errors from the system's prediction are to be attributed to the annotation process of the test set. For example, in the sentence "\textit{They included support for marine {\color{blue}\textit{reserves}} and money for fisheries management reform.}", \textit{reserves} is annotated as informal while the system identified it as formal.

In general, while bootstrapping the academic resource compilation and the informal word identification tasks, a minimal intervention of human annotators would enhance the overall system. Furthermore, integration of a BERT  or other contextualized embedding model \cite{devlin2018bert} could also help to improve the performance of the system.
Contextualized word embeddings provide word vector representations based on their context. As the vector representation of words varies as per the context, they implicitly provide a model for word sense disambiguation (WSD).

\section{Conclusion and Future Direction}
\label{conclusion}
In the realm of academic text writing, we explored how to compile academic resources, automatically identify informal words (words that are less formal for academic writing), and provide better substitutes. We have used a generic approach to compile the academic resources, which can be easily transferred to domains or languages as it only requires text corpus. The academic text rewriting system, analogous to lexical substitution systems, consists of informal word identification, candidate generation, candidate selection, and ranking components. As far as we know, this is the first experiment towards the development of academic writing support for academia, while there might be commercial cases (for example Grammarly\footnote{\url{https://www.grammarly.com/}}) that we do not know how the systems operate.

We envision this system to be embedded into open source academic writing aid tools where the academic sources are used to detect informal terms and propose academic substitutes. For the resource compilation process, it would be nice to extend the EmbedRank approach to extract keyphrases beyond the adjective and noun POS tag patterns, especially to cover verbs used in academic contexts.
% Furthermore, we plan to introduce techniques, that can generically build informal word identification resources beyond using a simple word list so that the system works even for technical and domain-dependent document composing (example medical domains). 

 Source code and resources of this the paper are released publicly\footnote{\url{https://github.com/uhh-lt/par4Acad}} on the Github repository under permissive licenses (ASL 2.0, CC-BY).

\section*{Acknowledgments}
This work was partially funded by a HKSAR UGC Teaching \& Learning Grant (Meeting the Challenge of Teaching and Learning Language in the University: Enhancing Linguistic Competence and Performance in English and Chinese) in the 2016-19 Triennium.

\nocite{*}
\section{Bibliographical References}
\label{main:ref}

\bibliographystyle{lrec}
\bibliography{lrec2020W-xample}

% \section{Language Resource References}
% \label{lr:ref}
% \bibliographystylelanguageresource{lrec}
% \bibliographylanguageresource{lrec2020W-xample}

\end{document}